\documentclass[runningheads]{llncs}
\usepackage[
    letterpaper,
    margin=1in
]{geometry}
\usepackage[T1]{fontenc}
\usepackage[
    colorlinks=true,
    linkcolor=blue,
    citecolor=blue,
    urlcolor=blue
]{hyperref}
\usepackage{newtxtext}
\usepackage{graphicx}
\usepackage{multirow}
\usepackage{amsfonts}
\usepackage{amsmath,amssymb,amsfonts}%
\usepackage{booktabs}%
\usepackage{array}
\usepackage{pifont}
\newcommand{\cmark}{\ding{51}}
\newcommand{\xmark}{\ding{55}}

\begin{document}
\title{Local-Frame Rotary Attention for Antibody-Specific Epitope Prediction}
\author{Chuanliu Fan\inst{1} \and
Junjie Wu\inst{1} \and
Nan Yu\inst{1} \and
Guohong Fu\inst{1,2}\textsuperscript{*}
}
\authorrunning{Fan et al.}
\institute{School of Computer Science and Technology, Soochow University \and
Institute of Artificial Intelligence, Soochow University\\
}
\maketitle
\begingroup
\renewcommand{\thefootnote}{\textasteriskcentered}
\footnotetext{Corresponding author. Email: ghfu@suda.edu.cn}
\endgroup

\begin{abstract}
Antibody-specific epitope prediction aims to identify which antigen residues are recognized by a given antibody, a task that depends on the three-dimensional complementarity between antibody CDRs and the antigen surface.
Existing methods usually leverage PLM embeddings and inject structure through additional graph, surface, or point-cloud encoders, where the positional mechanism inside attention remains largely tied to one-dimensional sequence order.
For proteins, the analogue of a token offset is not only sequence separation, but also the three-dimensional displacement between residues after folding.
This raises a question, 
can folded residue geometry serve as the positional mechanism of attention itself?
We propose Local-Frame Rotary Attention (LFRA), which expresses inter-residue displacements in backbone-defined local frames and injects them directly into rotary attention.
This design preserves continuous directional geometry while ensuring invariance to independent $\mathrm{SE}(3)$ transformations.
The code and dataset are available at \url{https://github.com/better-fcl/lfra}.

\keywords{Geometric rotary attention \and Antibody-specific epitope prediction \and $\mathrm{SE}(3)$ invariance.}

\end{abstract}

\section{Introduction}
Antibodies recognize antigens through specific sets of surface residues known as epitopes.
Identifying the epitope recognized by a given antibody is important for understanding antibody function, vaccine design, and therapeutic antibody engineering~\cite{pittala2020learning,del2021neural}.
Recent benchmarks such as AsEP~\cite{NEURIPS2024_15add673} have formalized this task and existing methods have made important progress by combining antibody information, protein language model (PLM) embeddings, and structural representations.
Early structure-based approaches such as EpiPred use manually designed geometric scoring over candidate surface patches~\cite{krawczyk2014improving}, while antigen-only models such as ESMBind and MaSIF-site identify generally interaction-prone regions from sequence or surface geometry~\cite{schreiber2023esmbind,gainza2020deciphering}.
WALLE and PEPNet, further improve performance by fusing antibody and antigen representations with graph or point-cloud geometry~\cite{NEURIPS2024_15add673,chen2026pepnet}.
However, these designs typically treat structure as an auxiliary representation processed by a separate graph, mesh, or point cloud encoder operating independently of the PLM embeddings.
In parallel, the positional mechanism inside attention of these methods often remains inherited from language modeling, where relative position is defined by one-dimensional sequence order.
This creates a mismatch for folded proteins.
Amino acid residues that are distant along the sequence can be adjacent in 3D space, and antibody recognition is governed by spatial proximity and orientation rather than by sequence distance alone.

This motivates a more direct question, can the geometry of the folded protein be used as the positional mechanism of attention itself? Rotary position embedding (RoPE) encodes relative token positions by rotating query and key subspaces according to their positional offsets~\cite{su2024roformer}.
For protein structures, the natural analogue of this offset is not only the sequence separation between two residues, but their three-dimensional displacement after folding.
Nevertheless, directly using displacements in a global Cartesian frame is not physically well-defined. 
The absolute axes of a PDB coordinate system are arbitrary, while epitope labels are scalar residue properties that should be invariant to global translations and rotations
% of the entire complex~\cite{berman2003announcing}. 
applied either jointly to the entire complex or independently to the antibody and antigen~\cite{berman2003announcing}.
Thus, an antibody-specific epitope predictor should retain continuous directional geometry while discarding nuisance variation from the global coordinate frame.

To this end, we propose Local-Frame 3D Rotary Attention (LFRA), a geometry attention mechanism for antibody-specific epitope prediction. For each residue, LFRA constructs a residue-centered backbone frame from its $\mathrm{N}$, $\mathrm{C}_{\alpha}$, and $\mathrm{C}$ atoms. The displacement from a query residue to a key residue is then expressed in the query residue's local frame and used to determine the rotary phase in attention. In this way, the model directly modulates residue--residue information propagation by local 3D geometry, rather than appending geometry as an ordinary node feature or discretizing it into distance-threshold edges.

We evaluate LFRA on the AsEP benchmark under both the epitope to antigen surface ratio split and the more challenging epitope group split.
LFRA achieves the best $\mathrm{MCC}$ on both splits, 
surpassing the strongest competing methods by $19.0\%$ and $22.4\%$, respectively.
Controlled comparisons with alternative geometry modules, including sequential RoFormer and global-coordinate SpaceFormer, further show that the gains come from the proposed local-frame geometric encoding rather than from simply injecting coordinates.
Transformation analyses verify that LFRA preserves residue-level predictions under whole-complex and independent antibody/antigen rigid transformations, while coordinate noise experiments demonstrate its robustness to residue-level positional perturbations.
Finally, antibody ablation experiments show that LFRA relies on antibody-specific information and the correct antibody--antigen pairing, rather than merely learning generic antigen-surface binding preferences.
Our main contributions are as follows:
\begin{itemize}
    \item We introduce LFRA, which encodes 3D displacements between residues via local backbone frames as rotary phases, addressing the mismatch between 1D positional encoding and the spatial proximity of folded proteins.
    \item LFRA achieves the best $\mathrm{MCC}$ on both splits of the AsEP benchmark and maintains robust predictions under whole-complex or independent $\mathrm{SE}(3)$ transformations, and residue-level coordinate perturbations.
    \item We demonstrate that LFRA learns antibody-conditioned epitope predictions that depend on the identity of the antibody and its correct pairing with the antigen, rather than relying solely on generic antigen-surface features, indicating its potential utility for protein interaction modeling.
\end{itemize}

\section{Related Work}
\subsection{Antibody Specific Epitope Prediction}
Early methods primarily rely on predefined structural representations~\cite{hamilton2017inductive,velickovic2017graph} and handcrafted scoring functions~\cite{jespersen2017bepipred}.
EpiPred~\cite{krawczyk2014improving} samples candidate surface patches from the antigen structure and ranks them using a graph-based scoring function defined over potential antibody--antigen residue pairs.
Sequence-driven structure predictors provide an alternative way to identify potential interfaces.
ESMFold~\cite{lin2023evolutionary}, which predicts protein structures from ESM-2 representations, can be used as a folding-based baseline by deriving interface residues from its predicted structures.
More recent approaches combine antibody and antigen information through learned sequence and structural representations.
WALLE~\cite{NEURIPS2024_15add673} adopts a multimodal framework that incorporates PLM representations together with structural features.
PEPNet~\cite{chen2026pepnet} represent protein surfaces using point cloud encoders.
These methods demonstrate the benefit of jointly modeling sequence, structure, and antibody context.
Nevertheless, structural information is generally processed by an additional graph, mesh, or point cloud encoder and subsequently fused with sequence representations.

A second class of methods predicts generic protein-binding regions using antigen information alone~\cite{wen2026patchepi}.
ESMBind~\cite{schreiber2023esmbind} fine-tunes ESM-2 with low-rank adaptation to predict binding-site residues directly from a single protein sequence.
MaSIF-site~\cite{gainza2020deciphering} instead represents the molecular surface as a mesh and applies geometric deep learning to classify surface vertices based on local geometric and physicochemical descriptors.
These methods can identify generally interaction-prone regions, but do not model which of these regions is recognized by a particular antibody.
Consequently, they cannot distinguish antibody-dependent epitopes on the same antigen.

\begin{figure*}[t]
\centering
\includegraphics[width=\textwidth]{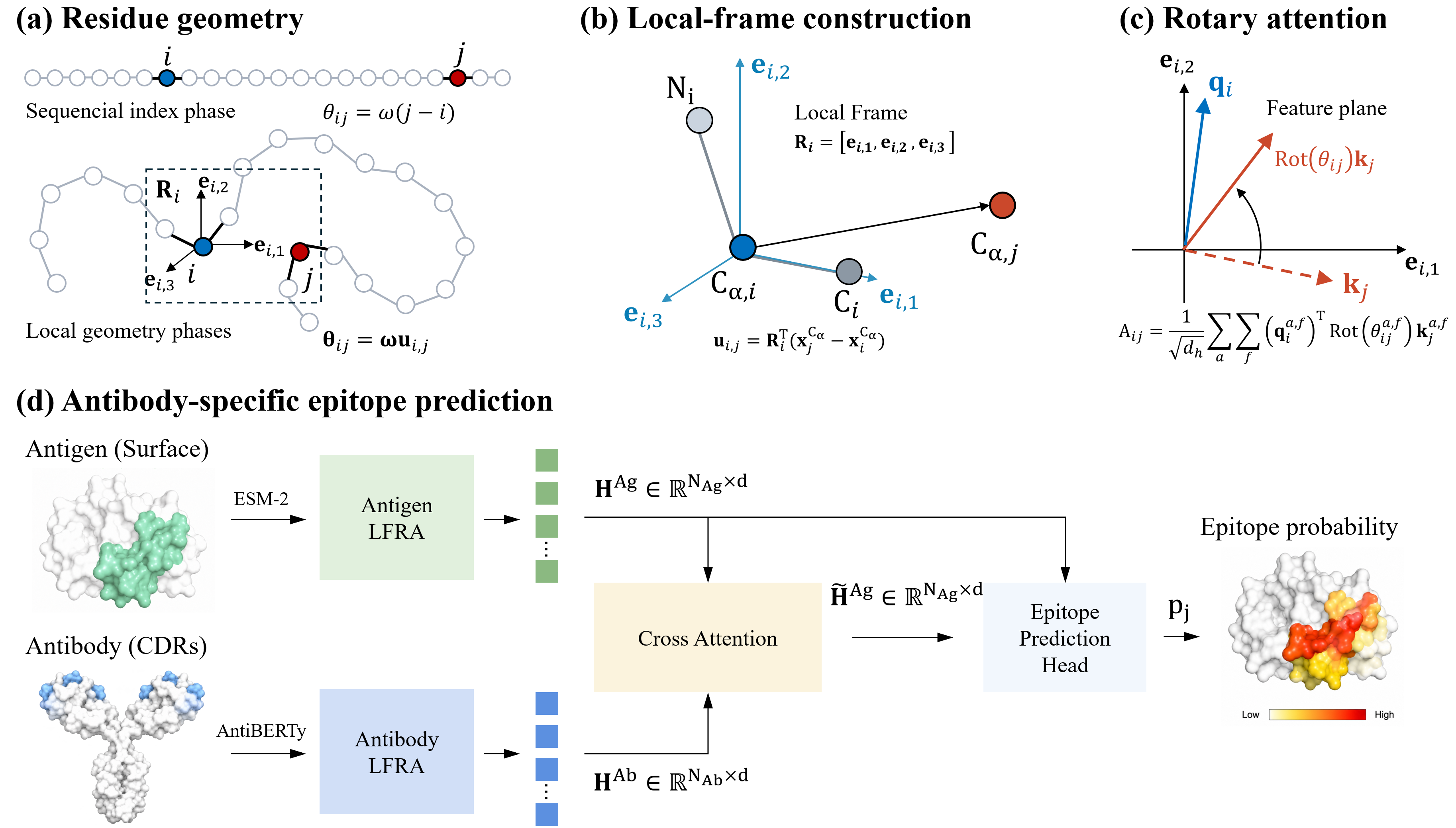}
\caption{
Overview of LFRA for antibody-specific epitope prediction.
(a) Comparison between conventional sequence-based rotary phase, derived from residue index differences, and local geometry-based phases, derived from pairwise displacements in residue-centered coordinate frames.
(b) Construction of a right-handed local backbone frame using the N, $\mathrm{C}_{\alpha}$, and C atoms, and representation of pairwise residue displacements in the query residue's local frame.
(c) Local-frame 3D rotary attention, where relative geometric displacements determine pair-specific rotations of key features before query--key interaction.
(d) Antibody-specific epitope prediction pipeline. Precomputed ESM-2 and AntiBERTy embeddings of antigen surface and antibody CDR residues, respectively, are processed by separate LFRA encoders. The resulting representations are integrated through antigen-to-antibody cross-attention, followed by a residue-level prediction head to estimate epitope probabilities on the antigen surface.
}
\label{fig1}
\end{figure*}

\subsection{Rotary Position Encoding}
Rotary position embedding (RoPE) incorporates positional information by applying position-dependent orthogonal rotations to two-dimensional query and key channel pairs~\cite{su2024roformer}.
Since the product of two rotary matrices depends only on the difference between their rotation angles, absolute token positions induce an explicit relative phase in the query--key inner product.
This multiplicative formulation preserves vector norms and avoids introducing a separate additive positional bias.
Subsequent studies have extended this principle beyond one-dimensional token indices. 
In multimodal models, M-RoPE~\cite{wang2024qwen2} and VideoRoPE~\cite{wei2025videorope} divide rotary channels among temporal and spatial axes.
More closely related to geometric modeling, SpaceFormer~\cite{lu2025beyond} extends RoPE to continuous three-dimensional coordinates by partitioning rotary channels among the global Cartesian axes.
Its query--key interaction therefore depends on the coordinate differences along the $x$, $y$, and $z$ axes.
These directional components depend on the arbitrary global orientation of the input coordinate system.
In contrast, our LFRA first expresses each inter-residue displacement in a residue-centered frame defined by the backbone atoms.
The frame co-rotates with the protein structure, making the resulting rotary phases invariant to global rotation and translation while retaining directional information relative to the local backbone geometry.

\section{Method}
\label{sec:method}

\subsection{Problem Formulation}
We formulate antibody-specific epitope prediction as a residue-level binary classification task.
Given an antibody--antigen complex, we retain the complementarity-determining region (CDR) residues of the antibody and the surface residues of the antigen, denoted by $\mathcal{V}_{\mathrm{Ab}}$ and $\mathcal{V}_{\mathrm{Ag}}$, respectively.
Following the definition adopted in AsEP~\cite{NEURIPS2024_15add673}, an antigen residue is labeled as an epitope if at least one of its non-hydrogen atoms lies within $4.5$~\AA{} of any non-hydrogen atom of the paired antibody.
The model jointly processes the antibody CDR residues and antigen surface residues and predicts an epitope probability for each antigen residue:
\begin{equation}
\hat{\mathbf{y}}=f_{\theta}\!\left(\mathcal{V}_{\mathrm{Ab}},\mathcal{V}_{\mathrm{Ag}}\right)\in [0,1]^{|\mathcal{V}_{\mathrm{Ag}}|}.
\label{eq:problem_formulation}
\end{equation}

\subsection{Local-Frame 3D Rotary Attention}
\label{sec:lf3drope}
Standard attention does not explicitly encode the three-dimensional relationship between residues.
Directly using Cartesian coordinates, however, makes the representation dependent on the arbitrary global orientation of the input structure.
We therefore introduce local-frame three-dimensional rotary position encoding, which injects relative residue geometry into the query--key interaction using residue-centered backbone frames.
The construction is applied separately within each partner, and we omit the partner index below.

For each residue $i$, let $\mathbf{x}_i^{\mathrm{N}}$, $\mathbf{x}_i^{\mathrm{C}_{\alpha}}$, and $\mathbf{x}_i^{\mathrm{C}}$ denote the coordinates of its $\mathrm{N}$, $\mathrm{C}_{\alpha}$, and $\mathrm{C}$ atoms, respectively.
We construct a right-handed local frame whose first axis follows the $\mathrm{C}_{\alpha}$-to-$\mathrm{C}$ direction:
\begin{equation}
\mathbf{e}_{i,1}=\frac{\mathbf{x}_i^{\mathrm{C}}-\mathbf{x}_i^{\mathrm{C}_{\alpha}}}{\left\|\mathbf{x}_i^{\mathrm{C}}-\mathbf{x}_i^{\mathrm{C}_{\alpha}}\right\|}.
\end{equation}
We remove the component parallel to $\mathbf{e}_{i,1}$ from the $\mathrm{C}_{\alpha}$-to-$\mathrm{N}$ direction:
\begin{equation}
\mathbf{v}_{i,2}=\mathbf{x}_i^{\mathrm{N}}-\mathbf{x}_i^{\mathrm{C}_{\alpha}}-\left[(\mathbf{x}_i^{\mathrm{N}}-\mathbf{x}_i^{\mathrm{C}_{\alpha}})^{\mathrm{T}}\mathbf{e}_{i,1}\right]\mathbf{e}_{i,1}.
\end{equation}
The remaining two axes are defined as
\begin{equation}
\mathbf{e}_{i,2}=\frac{\mathbf{v}_{i,2}}{\left\|\mathbf{v}_{i,2}\right\|},\qquad
\mathbf{e}_{i,3}=\mathbf{e}_{i,1}\times\mathbf{e}_{i,2}.
\end{equation}
As shown in Figure~\ref{fig1}, the local frame is represented by
\begin{equation}
\mathbf{R}_i=\left[\mathbf{e}_{i,1},\mathbf{e}_{i,2},\mathbf{e}_{i,3}\right].
\end{equation}

For a residue pair $(i,j)$, we use their $\mathrm{C}_{\alpha}$ atoms to represent the residue positions and express the displacement from residue $i$ to residue $j$ in the local frame of residue $i$:
\begin{equation}
\mathbf{u}_{ij}=\mathbf{R}_i^{\mathrm{T}}(\mathbf{x}_j^{\mathrm{C}_{\alpha}}-\mathbf{x}_i^{\mathrm{C}_{\alpha}})
=(u_{ij,1},u_{ij,2},u_{ij,3})^{\mathrm{T}}.
\label{eq:local_pair_vector}
\end{equation}
Thus, $u_{ij,a}$ describes the relative displacement along the $a$-th local backbone axis.
Unlike scalar distances, $\mathbf{u}_{ij}$ preserves directional information in a residue-centered coordinate system.
For residues with missing or degenerate backbone coordinates, we set the corresponding local displacement to zero, making the rotary transformation the identity.

We next use $\mathbf{u}_{ij}$ to modulate self-attention.
Let $H$ be the number of attention heads and $d_h=d/H$ the dimension of each head.
The rotary channels of each head are divided into three groups, corresponding to the three local axes.
Each group contains $F$ two-dimensional channel pairs associated with spatial frequencies $\{\omega_f\}_{f=1}^{F}$.
The remaining $r=d_h-6F$ channels are left unrotated.
For local axis index $a$ and frequency $f$, the pair-specific rotation angle is
\begin{equation}
\theta_{ij}^{a,f}=\omega_f u_{ij,a}.
\label{eq:lf3drope_angle}
\end{equation}
Different frequencies provide multiple resolutions of the same relative displacement, allowing the model to capture both coarse and fine geometric relationships.
Let $\mathbf{q}_{i}^{a,f},\mathbf{k}_{j}^{a,f}\in\mathbb{R}^{2}$ denote the corresponding query and key channel pairs within a single attention head.
We rotate the key pair using
\begin{equation}
\mathrm{Rot}(\theta)=
\left[\begin{array}{cc}
\cos\theta & -\sin\theta\\
\sin\theta & \cos\theta
\end{array}\right].
\end{equation}
The geometry-aware attention logit is then
\begin{equation}
\mathrm{A}_{ij}=\frac{1}{\sqrt{d_h}}
\sum_{a=1}^{3}\sum_{f=1}^{F}
(\mathbf{q}_{i}^{a,f})^{\mathrm{T}}
\mathrm{Rot}(\theta_{ij}^{a,f})
\mathbf{k}_{j}^{a,f}.
\label{eq:lf3drope_attention}
\end{equation}
The remaining unrotated channels contribute through the standard query--key dot product $\mathbf{q}_{i}^{\mathrm{T}}\mathbf{k}_{j}$.
The logits are normalized over valid key residues using softmax, after which the value vectors are aggregated as in standard multi-head self-attention~\cite{NIPS2017_3f5ee243}.
The head outputs are concatenated and linearly projected back to the model dimension.
Therefore, the local relative position of each residue pair directly modulates its query--key compatibility.
The design principles and detailed selection of rotary frequencies are provided in Appendix~\ref{sec:ladder}.

\subsection{Geometry Encoding and Antibody Conditioning}
\label{sec:antibody_specific_prediction}

For each partner $M\in\{\mathrm{Ab},\mathrm{Ag}\}$, let $\mathbf{E}^{M}\in\mathbb{R}^{N_M\times p_M}$ denote the precomputed protein language model (PLM) embeddings of its selected residues, where $N_M$ is the number of residues and $p_M$ is the embedding dimension.
We use separate learnable projection modules to map the two embedding spaces into a common $d$-dimensional representation:
\begin{equation}
\mathbf{H}^{M}_0=\phi_M(\mathbf{E}^{M})\in\mathbb{R}^{N_M\times d},
\qquad M\in\{\mathrm{Ab},\mathrm{Ag}\},
\label{eq_h0}
\end{equation}

The antibody and antigen are independently processed by single-layer geometry encoders with identical architectures but separate parameters.
Each encoder applies the proposed local-frame rotary attention to its partner-specific residue features and backbone coordinates, producing geometry-aware representations $\mathbf{H}^{\mathrm{Ab}}$ and $\mathbf{H}^{\mathrm{Ag}}$, respectively.
To incorporate antibody-specific context into antigen representations, we apply two successive antigen-to-antibody cross-attention layers:
\begin{equation}
\mathbf{H}^{\mathrm{Ag|Ab}}
=
\operatorname{Cross}
(\mathbf{H}^{\mathrm{Ag}},\mathbf{H}^{\mathrm{Ab}}).
\label{eq:cross_attention}
\end{equation}

In each layer, antigen representations provide queries, while the independently encoded antibody representations provide keys and values.
Importantly, cross-attention relies solely on learned feature interactions and does not use antibody--antigen coordinate differences.
Together with the independent local-frame geometry encoding, this design makes the prediction invariant to independent rigid transformations of the two partners and eliminates the need for their relative binding pose.
Finally, we predict epitope probabilities at the antigen residue level.
Let $\mathbf{h}^{\mathrm{Ag}}_j$ and $\mathbf{h}^{\mathrm{Ag|Ab}}_j$ denote the geometry-encoded and antibody-conditioned representations of antigen residue $j$, respectively.
We combine these representations through concatenation and element-wise interaction:
\begin{equation}
\begin{aligned}
\mathbf{z}_j
&=\mathbf{h}^{\mathrm{Ag}}_j
\Vert\mathbf{h}^{\mathrm{Ag|Ab}}_j
\Vert
(\mathbf{h}^{\mathrm{Ag}}_j\odot\mathbf{h}^{\mathrm{Ag|Ab}}_j),\\
s_j&=\phi_{\mathrm{out}}(\mathbf{z}_j),
\qquad
p_j=\sigma(s_j),
\end{aligned}
\label{eq:epitope_prediction}
\end{equation}
where $\odot$ denotes element-wise multiplication, $\Vert$ denotes concatenation, and $\phi_{\mathrm{out}}$ is a two-layer perceptron with hidden width $d$.
The sigmoid function $\sigma$ converts the resulting logit $s_j$ into the epitope probability $p_j$.

\section{Experiments}
\subsection{Experimental Setup}

\subsubsection{Dataset}
We conduct experiments on AsEP~\cite{NEURIPS2024_15add673}, a residue-level antibody-specific epitope prediction benchmark containing 1,723 non-redundant antibody--antigen complexes curated from Antibody Database~\cite{ferdous2018abdb}. Following the preprocessing of AsEP and PEPNet~\cite{chen2026pepnet}, each complex is represented by antibody CDR residues and antigen surface residues, and the task is formulated as binary classification over antigen surface residues. We directly use the official split lists provided by PEPNet. Specifically, we consider the \emph{epitope to antigen ratio split} and the more challenging \emph{epitope group split}, both of which contain 1,383 training, 170 validation, and 170 test complexes. The former maintains comparable epitope-to-antigen-surface residue ratios across the three subsets, whereas the latter ensures that test-set epitope groups are absent from both the training and validation sets, thereby evaluating generalization to unseen binding regions.

\subsubsection{Baselines}
\begin{table*}[t]
\centering
\caption{
Summary of features used in benchmarking methods.
$^\dagger$PEPNet additionally use local surface normal vectors, position-specific scoring matrices, solvent accessibility, and neighbor composition.
Antibody indicates whether antibody information is considered when predicting epitope residues;
Amino acid and Atom denote whether explicit residue and atom features are used. 
}
\small
\begin{tabular}{@{}lcccccc@{}}
\toprule
Method& Antibody& Structure& \shortstack{PLM(Antigen / Antibody)}& \shortstack{Amino acid}& \shortstack{Atom}& Other \\
\midrule
EpiPred& \cmark& Graph& \xmark& \xmark& \xmark& \xmark \\
ESMFold& \cmark& \xmark& ESM-2 / ESM-2& \xmark& \xmark& \xmark \\
MaSIF-site& \xmark& Graph& \xmark& \xmark& \xmark& \xmark \\
ESMBind& \xmark& \xmark& ESM-2 / --& \xmark& \xmark& \xmark \\
WALLE& \cmark& Graph& \shortstack{ESM-2 / AntiBERTy, IgFold}& \xmark& \xmark& \xmark \\
PEPNet\textsuperscript{$\dagger$}& \cmark& PointNet& \xmark& \cmark& one-hot& \cmark \\
PEPNet+LE\textsuperscript{$\dagger$}& \cmark& PointNet& ESM-2 / AntiBERTy& \cmark& one-hot& \cmark \\
\midrule
LFRA& \cmark& \shortstack{Local-frame Rotary Attention}& ESM-2 / AntiBERTy& \xmark& \xmark& \xmark \\
\bottomrule
\end{tabular}
\label{tab:feature_summary}
\end{table*}

\begin{table*}[th]
\centering
\caption{
Main results on the AsEP benchmark under
(a) the epitope to antigen surface ratio split and
(b) the epitope group split.
Given the substantial class imbalance in epitope prediction,
$\mathrm{MCC}$ is regarded as the primary evaluation metric.
LFRA results are reported as mean $\pm$ standard deviation over five runs.
The best and second-best results for each metric are highlighted in
bold and underlined, respectively.
}
\small
\setlength{\tabcolsep}{6pt}
\begin{tabular}{@{}lccccc@{}}
\multicolumn{6}{c}{
    (a) Epitope to antigen surface ratio split
} \\
\toprule
Method & $\mathrm{MCC}$ & $\mathrm{Precision}$ & $\mathrm{Recall}$ & $\mathrm{AUROC}$ & $\mathrm{F1}$ \\
\midrule

ESMFold
& $0.028\pm0.010$
& $0.137\pm0.019$
& $0.043\pm0.006$
& --
& $0.060\pm0.008$ \\

ESMBind
& $0.016\pm0.008$
& $0.106\pm0.012$
& $0.121\pm0.014$
& $0.506\pm0.004$
& $0.090\pm0.009$ \\

EpiPred
& $0.029\pm0.018$
& $0.122\pm0.014$
& $0.180\pm0.019$
& --
& $0.142\pm0.016$ \\

MaSIF-site
& $0.037\pm0.012$
& $0.125\pm0.015$
& $0.183\pm0.017$
& --
& $0.114\pm0.011$ \\

WALLE
& $0.210\pm0.020$
& $0.235\pm0.018$
& $\mathbf{0.422\pm0.028}$
& $0.635\pm0.013$
& $0.258\pm0.018$ \\

PEPNet
& $\underline{0.401}$
& $\underline{0.544}$
& $0.340$
& $\underline{0.765}$
& $\underline{0.419}$ \\
PEPNet+LE
& $0.337$
& $0.465$
& $0.295$
& $\underline{0.765}$
& $0.361$ \\
\midrule
LFRA
& $\mathbf{0.477\pm0.010}$
& $\mathbf{0.675\pm0.040}$
& $\underline{0.374\pm0.018}$
& $\mathbf{0.846\pm0.002}$
& $\mathbf{0.480\pm0.009}$ \\
\bottomrule
\end{tabular}

\begin{tabular}{@{}lccccc@{}}
\multicolumn{6}{c}{
    (b) Epitope group split
} \\
\toprule
Method & $\mathrm{MCC}$ & $\mathrm{Precision}$ & $\mathrm{Recall}$ & $\mathrm{AUROC}$ & $\mathrm{F1}$ \\
\midrule

ESMFold
& $0.018\pm0.010$
& $0.113\pm0.019$
& $0.034\pm0.007$
& --
& $0.046\pm0.009$ \\

ESMBind
& $0.002\pm0.008$
& $0.082\pm0.011$
& $0.076\pm0.011$
& $0.500\pm0.004$
& $0.064\pm0.008$ \\

EpiPred
& $-0.006\pm0.015$
& $0.089\pm0.011$
& $0.158\pm0.019$
& --
& $0.112\pm0.014$ \\

MaSIF-site
& $0.046\pm0.014$
& $0.164\pm0.020$
& $\underline{0.174\pm0.015}$
& --
& $0.128\pm0.012$ \\

WALLE
& $0.077\pm0.015$
& $0.143\pm0.017$
& $\mathbf{0.266\pm0.025}$
& $0.544\pm0.010$
& $0.145\pm0.014$ \\

PEPNet
& $0.139$
& $0.230$
& $0.143$
& $0.612$
& $0.177$ \\

PEPNet+LE
& $\underline{0.156}$
& $\underline{0.250}$
& $0.155$
& $\underline{0.627}$
& $\underline{0.191}$ \\

\midrule

LFRA
& $\mathbf{0.191\pm0.015}$
& $\mathbf{0.323\pm0.027}$
& $0.160\pm0.014$
& $\mathbf{0.691\pm0.012}$
& $\mathbf{0.213\pm0.014}$ \\
\bottomrule
\end{tabular}
\label{tab:main_ratio_groups}
\end{table*}

To evaluate our method, we compare it with seven representative baselines spanning different modeling paradigms: two sequence-based methods, ESMFold~\cite{lin2023evolutionary} and ESMBind~\cite{schreiber2023esmbind}; two structure-based methods, EpiPred~\cite{krawczyk2014improving} and MaSIF-site~\cite{gainza2020deciphering}; one multimodal method, WALLE~\cite{NEURIPS2024_15add673}; and two point-cloud-based methods, PEPNet and PEPNet+LE~\cite{chen2026pepnet}.
Their input features are summarized in Table~\ref{tab:feature_summary}.
To isolate the effect of geometric positional encoding, we further evaluate nine matched architectural variants as shown in Table~\ref{tab:geometry_injection_comparison} in section~\ref{sec:geometry-modules}.
For example, RoFormer uses conventional sequence-index rotary position encoding without explicit 3D geometric terms. 
SpaceFormer instead uses inter-residue displacement vectors expressed in the global input coordinate frame.
Except for the geometry module, the input features and the remaining modules are kept unchanged across these variants.

\subsubsection{Metrics}
\label{sec:metrics}
We use the Matthews correlation coefficient ($\mathrm{MCC}$) as the primary evaluation metric, following prior recommendations \cite{matthews1975comparison,chicco2020advantages}.
$\mathrm{MCC}$ accounts for all four outcomes in the confusion matrix and provides a balanced assessment under the substantial class imbalance between epitope and non-epitope residues. 
It is considered superior to the $\mathrm{AUROC}$, which is evaluated over all thresholds~\cite{NEURIPS2024_15add673}.
For completeness and comparison with prior work, we also report area under the receiver operating characteristic curve ($\mathrm{AUROC}$), $\mathrm{Precision}$, $\mathrm{Recall}$, and $\mathrm{F1}$-score. 
All threshold-dependent metrics are computed using a probability threshold selected on the validation set to maximize $\mathrm{MCC}$ and then fixed for test set evaluation~\cite{chen2026pepnet}.

\subsubsection{Implementation Details}
\label{sec:implementation_details}
We implement LFRA in PyTorch and conduct all experiments on a single NVIDIA A800 GPU using FP32 precision.
The model takes AsEP's precomputed AntiBERTy embeddings for antibody residues, extracted via IgFold~\cite{ruffolo2021deciphering,ruffolo2023fast}, and ESM-2 embeddings~\cite{lin2023evolutionary} for antigen residues, together with the N, $\mathrm{C}_{\alpha}$, and C coordinates of the selected CDR and antigen surface residues.
The PLM embeddings remain fixed during training.
To address residue-level class imbalance, we jointly optimize all trainable components using weighted binary cross-entropy combined with a soft Dice loss~\cite{milletari2016v}.
We use five random seeds, and results are reported as the mean $\pm$ sample standard deviation.
The complete configurations are summarized in Table~\ref{tab:implementation_hyperparameters} of Appendix~\ref{sec:appendix_implementation_details}.

\subsection{Main Results}
Table~\ref{tab:main_ratio_groups} presents the performance of LFRA and existing methods on the two official AsEP splits.
Since epitope prediction is highly class-imbalanced, we regard $\mathrm{MCC}$ as the primary metric. LFRA achieves the best $\mathrm{MCC}$ on both splits, reaching $0.477\pm0.010$ on the ratio split and $0.191\pm0.015$ on the more challenging group split. On the ratio split, LFRA also obtains the best $\mathrm{AUROC}$, precision, and $\mathrm{F1}$, outperforming the strongest prior $\mathrm{MCC}$ baseline PEPNet while keeping a more balanced precision--recall trade-off than WALLE. On the group split, LFRA further improves over both PEPNet and PEPNet+LE in $\mathrm{MCC}$ and $\mathrm{F1}$, indicating better generalization to unseen epitope groups.
These results demonstrate the effectiveness and robustness of LFRA across different data partitioning strategies.

\section{Analysis}
\subsection{Equipped with Other Geometry Modules}
\label{sec:geometry-modules}

\begin{table*}[t]
\centering
\caption{Comparison of geometry injection methods. All variants share the PLM
inputs, input projections, cross-attention and prediction head. Structural
features are derived solely from AsEP structures.}
\small
\begin{tabular}{lll}
\toprule
Method & Geometric features & Injection mechanism \\
\midrule

NoGeometry
& --
& No residue encoder \\

RoFormer
& Selected-residue indices
& 1D rotary attention \\

SpaceFormer
& Displacements in global coordinates
& 3D rotary attention \\

DistanceRBF
& Gaussian RBFs of $\mathrm{C}_{\alpha}$ distances
& Additive attention bias \\

LFGB
& Distance RBFs, local unit directions and relative frame orientations
& Additive attention bias \\

IPA
& Feature-predicted points in backbone frames
& Point-distance bias and point aggregation \\

GVP
& Backbone vectors, distance RBFs and edge directions
& Scalar/vector message passing \\

Geformer
& Exponential-normal distance RBFs with a cosine cutoff
& Multiplicative attention on a radius graph \\

DeepInteract
& Distances, local directions, orientations, torsions, plane angles
& Edge-conditioned attention on a $k$NN graph \\

\midrule
LFRA
& Displacements in backbone local frames
& 3D rotary attention \\
\bottomrule
\end{tabular}
\label{tab:geometry_injection_comparison}
\end{table*}

\begin{figure*}[t]
    \centering
    \includegraphics[width=\textwidth]{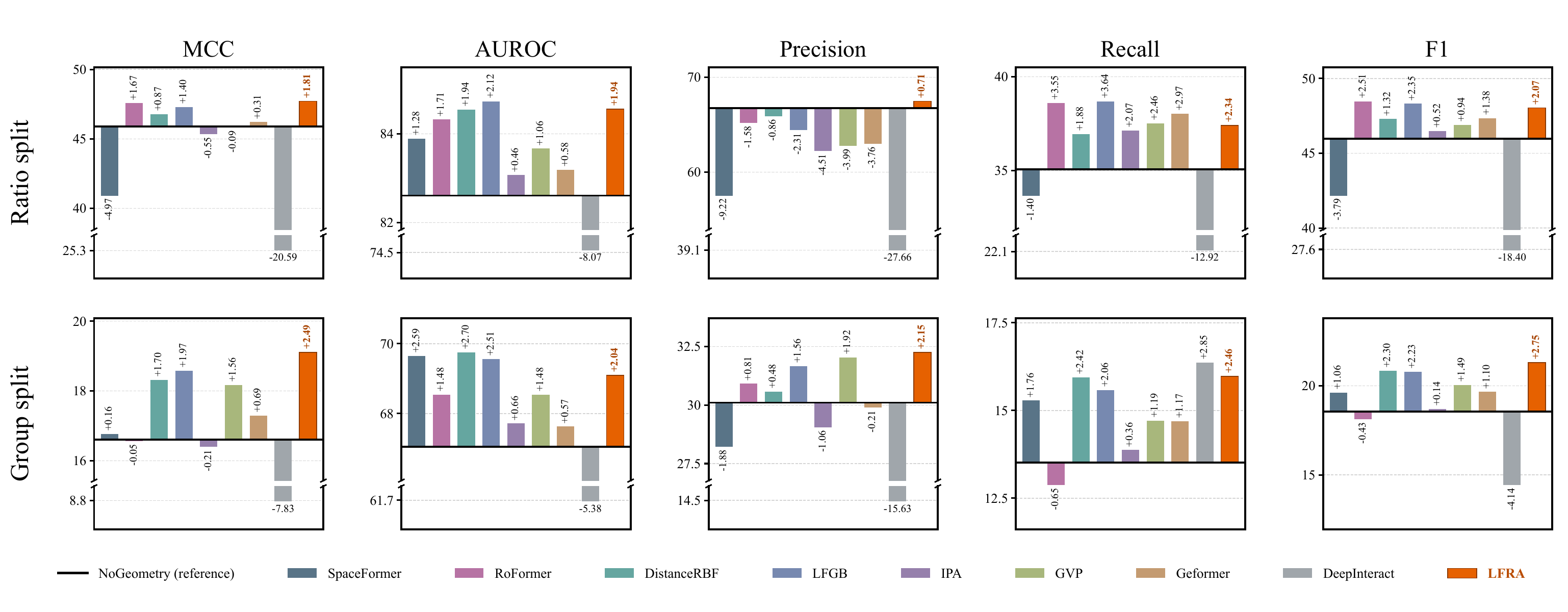}
    \caption{Performance changes of different geometry modules relative to NoGeometry on the two official AsEP splits, averaged over five random seeds.
    Positive values indicate improvements, while negative values indicate performance degradation.}
    \label{fig:geometry_modules}
\end{figure*}

Table~\ref{tab:geometry_injection_comparison} summarizes the geometric representations and injection mechanisms of the evaluated methods.
To assess the contribution of geometric encoding, we compare LFRA with eight alternatives while keeping the PLM inputs, input projections, cross-attention, and prediction head fixed. NoGeometry serves as the reference baseline, removing the residue-level geometry encoder.
All structural features are derived solely from the AsEP structures.
Each result is averaged over five runs, and Figure~\ref{fig:geometry_modules} reports the differences in MCC, AUROC, precision, recall, and F1 relative to NoGeometry on the official AsEP ratio and group splits.
The compared methods cover a range of geometric representations.
RoFormer~\cite{su2024roformer} applies rotary positional embeddings to selected-residue indices, SpaceFormer~\cite{lu2025beyond} extends rotary attention to displacements in global coordinates.
DistanceRBF encodes pairwise C$_\alpha$ distances using Gaussian radial basis functions (RBFs) as additive attention biases.
Local-Frame Geometric Bias( LFGB)~\cite{ganea2021independent} additionally incorporates local pair directions and relative frame orientations.
IPA~\cite{jumper2021highly} represents geometric relationships through learned points anchored to residue frames.
GVP~\cite{jing2021learning} propagates scalar and equivariant vector features along structural edges.
Geformer~\cite{wang2023geometric} employs exponential-normal distance RBFs with a cosine cutoff on a radius graph, while DeepInteract~\cite{deepinteract} uses distances, local directions, orientations, torsions, and plane angles to construct edge-conditioned attention on a $k$-nearest-neighbor graph.
In contrast, LFRA encodes pairwise displacements in residue-local backbone frames and incorporates them through 3D rotary attention.

As shown in Figure~\ref{fig:geometry_modules}, LFRA consistently improves upon NoGeometry across all five metrics on both splits, with particularly strong gains in MCC and F1 on the group split.
Although several alternatives achieve competitive results on individual metrics, their improvements are less consistent across the two splits.
For example, DistanceRBF and LFGB yield larger AUROC gains on the ratio split, whereas GVP improves MCC on the ratio split but does not maintain this advantage on the group split. DeepInteract also exhibits a precision--recall trade-off, with improved recall but substantially reduced precision on the group split. 
The comparison between SpaceFormer and LFRA further suggests that encoding pairwise geometry in local backbone frames is more effective than directly using global-coordinate displacements.
Overall, LFRA achieves the largest MCC gains on both splits and the largest F1 gain on the group split, indicating more consistent improvements.

\begin{figure*}[t]
    \centering
    \includegraphics[width=0.9\textwidth]{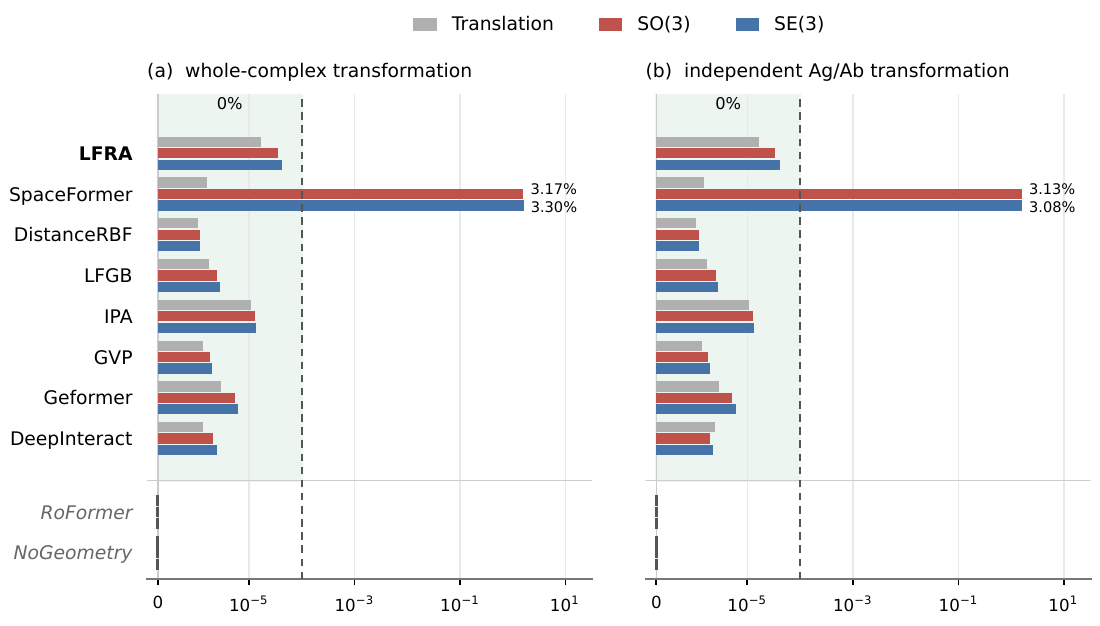}
    \caption{Model sensitivity to rigid transformations under (a) whole-complex and (b) independent Ag/Ab transformations.
    Bars indicate mean absolute logit changes on a symmetric-log scale.
    Logit changes are considered negligible below the dashed line.
    Percentages denote the fraction of antigen surface residues whose binary epitope predictions change after transformation.}
    \label{fig:transformation}
\end{figure*}

\subsection{SE(3) Transformation Analysis}
\label{sec:analysis-rigid}

We evaluate model sensitivity under whole-complex and independent antigen/antibody rigid transformations.
The sampling and transformation protocols are detailed in Table~\ref{tab:implementation_hyperparameters}.
Figure~\ref{fig:transformation} reports the mean absolute logit changes and epitope decision-flip rates.
The epitope decision-flip rates measure the percentage of antigen surface residues whose binary predictions change after transformation.
Results are averaged across the ratio and group splits.

In Panel~(a), LFRA exhibits the expected invariance to global rigid transformations.
Under $\mathbf{x}\mapsto\mathbf{Q}\mathbf{x}+\mathbf{t}$, where $\mathbf{Q}\in\mathrm{SO}(3)$, the local frame transforms as $\mathbf{R}_i\mapsto\mathbf{Q}\mathbf{R}_i$, leaving the local displacement $\mathbf{R}_i^{\mathrm{T}}(\mathbf{x}_j-\mathbf{x}_i)$ unchanged.
Consequently, LFRA produces zero decision flips across both splits, with a mean absolute logit change of only $3.1\times10^{-5}$.
The six other geometry-based methods, except SpaceFormer, also preserve all residue-level decisions.
RoFormer and NoGeometry produce exactly zero logit changes, as expected for methods without coordinate-dependent geometric encoding.
In contrast, SpaceFormer is invariant to translation but sensitive to rotation, exhibiting decision-flip rates of $3.17\%$ and $3.30\%$ under $\mathrm{SO}(3)$ and $\mathrm{SE}(3)$ transformations, respectively, with a mean absolute logit change of approximately $1.6$.
Panel~(b) examines whether predictions depend on the relative spatial arrangement of the antibody and antigen.
Independently transforming Ag/Ab changes their relative binding pose while preserving their respective internal geometries.
LFRA and the six other geometry-based methods, as well as RoFormer and NoGeometry, retain all residue-level decisions under these transformations.
These results indicate that the evaluated architectures do not explicitly encode inter-partner spatial relationships.
SpaceFormer again exhibits sensitivity to rotation, with decision-flip rates of $3.13\%$ and $3.08\%$ under independent $\mathrm{SO}(3)$ and $\mathrm{SE}(3)$ transformations, respectively, and a mean absolute logit change of approximately $1.6$.
Its comparable sensitivity across the two protocols suggests that the prediction changes arise from dependence on global coordinate orientations rather than the relative binding pose itself.
Overall, LFRA preserves both residue-level logits and epitope predictions under global rigid transformations and independent transformations of the two partners, demonstrating its invariance without requiring the relative binding geometry.

\begin{figure*}[t]
    \centering
    \includegraphics[width=\textwidth]{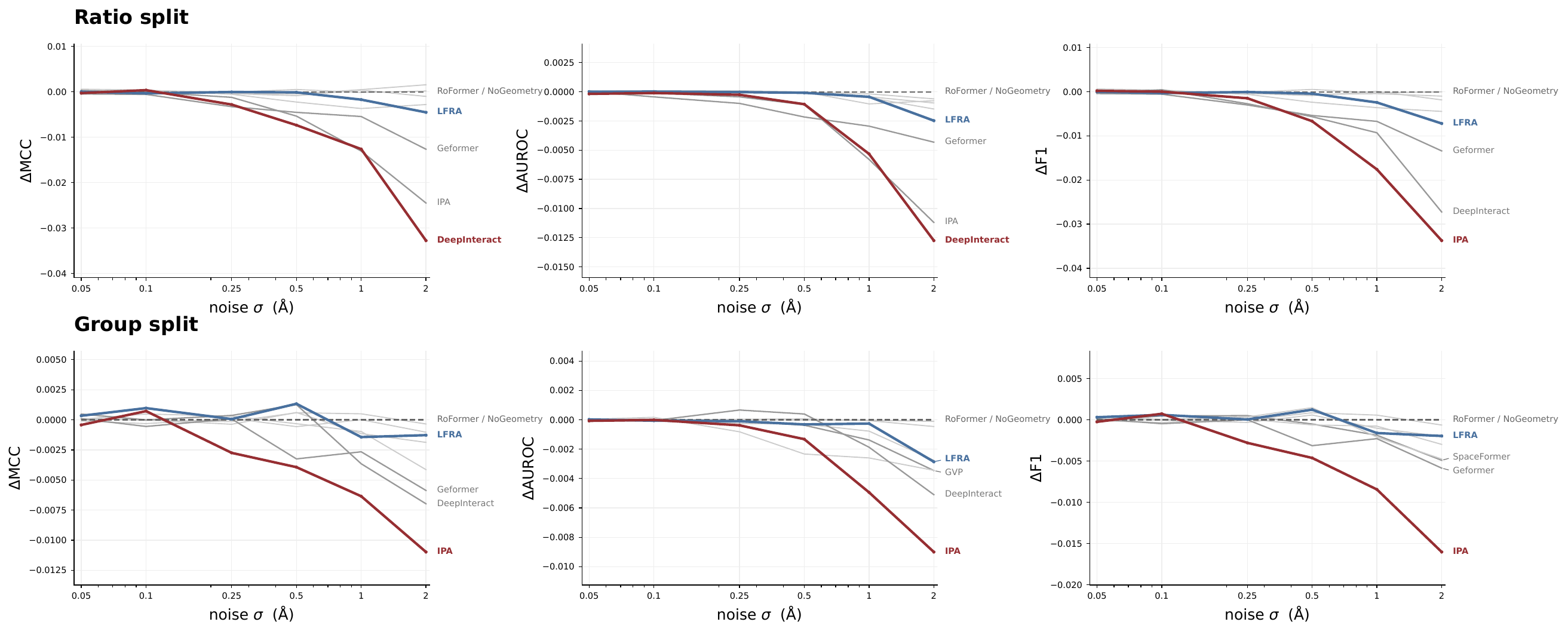}
    \caption{Performance changes under residue-level coordinate noise on the official AsEP ratio and group splits.
    Changes in MCC, AUROC, and F1 are reported relative to predictions on the unperturbed structures.}
    \label{fig:noise}
\end{figure*}

\subsection{Coordinate Noise Analysis}
\label{sec:analysis-noise}

We evaluate the robustness of different geometry modules to coordinate uncertainty by adding isotropic Gaussian noise at the residue level.
For each residue $i$, we independently sample a displacement $\boldsymbol{\epsilon}_i\sim\mathcal{N}(\mathbf{0},\sigma^2\mathbf{I})$ and apply the same displacement to its N, $\mathrm{C}_{\alpha}$, and C atoms.
This residue-wise perturbation preserves the internal backbone geometry of each residue while changing its spatial position, avoiding the nonphysical bond and angle distortions caused by independently perturbing individual atoms.
We evaluate six noise levels, $\sigma\in\{0.05,0.1,0.25,0.5,1.0,2.0\}$~\AA{}, on the full test set of 170 complexes, covering 46,395 antigen surface residues in the ratio split and 45,193 in the group split.
Figure~\ref{fig:noise} reports the changes in MCC, AUROC, and F1 relative to predictions on the unperturbed structures.

As shown in Figure~\ref{fig:noise}, most geometry-based methods exhibit increasingly degraded performance as the noise level increases, although their sensitivity varies across methods and splits.
LFRA maintains relatively stable performance under residue-level perturbations, with particularly small changes on the group split.
On the ratio split, its performance also degrades at higher noise levels, but the reduction remains considerably smaller than that of more sensitive methods such as IPA and DeepInteract.
The relatively large performance drops of these methods suggest greater sensitivity to perturbations in residue positions.
Although the relative displacements between residues are still affected, the local reference frames remain stable, avoiding additional errors caused by perturbing the backbone orientations.
These results indicate that LFRA provides favorable robustness to residue-level coordinate uncertainty.

\begin{table*}[t]
\centering
\caption{Antibody-ablation results (mean $\pm$ standard deviation over five seeds). Removing or
replacing the antibody consistently degrades the full LFRA model. The \emph{retrained}
row is an antigen-only model trained from scratch; the remaining rows ablate the
antibody of the trained model at inference time.}
\label{tab:antibody_ablation}
\small
\begin{tabular}{lccccc}
\toprule
Method & $\mathrm{MCC}$ & $\mathrm{Precision}$ & $\mathrm{Recall}$ & $\mathrm{AUROC}$ & $\mathrm{F1}$ \\
\midrule
\multicolumn{6}{l}{\textit{Group split}} \\
\midrule
\textbf{LFRA (full)} & $\mathbf{0.191\pm0.015}$ & $\mathbf{0.323\pm0.027}$ & $\mathbf{0.160\pm0.014}$ & $\mathbf{0.691\pm0.012}$ & $\mathbf{0.213\pm0.014}$ \\
LFRA w/o Ab (retrained) & $0.139\pm0.016$ & $0.229\pm0.015$ & $0.147\pm0.030$ & $0.675\pm0.006$ & $0.177\pm0.022$ \\
LFRA w/o Ab (zero) & $0.116\pm0.025$ & $0.210\pm0.030$ & $0.119\pm0.020$ & $0.669\pm0.020$ & $0.151\pm0.023$ \\
LFRA w/o Ab (mean) & $0.103\pm0.021$ & $0.198\pm0.035$ & $0.107\pm0.017$ & $0.681\pm0.018$ & $0.138\pm0.018$ \\
LFRA w/o Ab (mismatched) & $0.107\pm0.018$ & $0.217\pm0.018$ & $0.097\pm0.019$ & $0.666\pm0.019$ & $0.133\pm0.021$ \\
\midrule
\multicolumn{6}{l}{\textit{Ratio split}} \\
\midrule
\textbf{LFRA (full)} & $\mathbf{0.477\pm0.010}$ & $\mathbf{0.675\pm0.040}$ & $\mathbf{0.374\pm0.018}$ & $\mathbf{0.846\pm0.002}$ & $\mathbf{0.480\pm0.009}$ \\
LFRA w/o Ab (retrained) & $0.319\pm0.017$ & $0.296\pm0.025$ & $0.479\pm0.057$ & $0.834\pm0.004$ & $0.364\pm0.014$ \\
LFRA w/o Ab (zero) & $0.190\pm0.044$ & $0.376\pm0.055$ & $0.131\pm0.044$ & $0.789\pm0.026$ & $0.192\pm0.052$ \\
LFRA w/o Ab (mean) & $0.181\pm0.027$ & $0.378\pm0.026$ & $0.118\pm0.032$ & $0.800\pm0.014$ & $0.178\pm0.036$ \\
LFRA w/o Ab (mismatched) & $0.180\pm0.029$ & $0.357\pm0.037$ & $0.125\pm0.027$ & $0.779\pm0.007$ & $0.185\pm0.032$ \\
\bottomrule
\end{tabular}
\end{table*}

\subsection{Is the Model Antibody-Specific?}
\label{sec:antibody_ablation}

An antibody-specific epitope predictor should condition its predictions on the identity of the antibody rather than produce a generic binding map of the antigen surface.
To examine this, we compare the full LFRA model with four antibody ablation settings, as summarized in Table~\ref{tab:antibody_ablation}.
First, \textit{w/o Ab (retrained)} removes the antibody branch entirely and retrains the resulting antigen-only model from scratch using the same architecture and training configuration.
The remaining settings ablate the antibody at inference time using the trained full model: \textit{w/o Ab (zero)} replaces the antibody sequence features and backbone coordinates with zeros; \textit{w/o Ab (mean)} replaces its sequence features with the mean antibody embedding computed over the training split; and \textit{w/o Ab (mismatched)} supplies a real antibody from a different complex while retaining the original antigen.
These settings distinguish the effect of removing antibody information from that of providing an incorrect antibody, while the retrained model controls for potential distribution shifts caused by inference-time ablation.

As shown in Table~\ref{tab:antibody_ablation}, removing or replacing the antibody consistently degrades performance on both splits.
On the ratio split, MCC decreases from $0.477$ for the full model to $0.319$ for the retrained antigen-only model and further to approximately $0.19$ for inference-time antibody ablations.
A similar trend is observed on the group split, where MCC decreases from $0.191$ to $0.139$ and further to approximately $0.12$.
Two observations are particularly important.
First, although the antigen-only model retains some predictive capability, its performance remains substantially below that of the full model, indicating that antibody information provides essential context for epitope prediction.
Second, replacing the correct antibody with a real but mismatched antibody yields little improvement over removing the antibody, suggesting that the model relies on the correspondence between the antibody and antigen rather than merely the presence of an antibody input.
This finding also indicates that the performance degradation cannot be explained solely by the out-of-distribution nature of zero or mean antibody inputs.
Together, these results demonstrate that LFRA learns antibody-conditioned epitope predictions rather than generic antigen-surface binding preferences.

\section{Conclusion}
We presented LFRA, a local-frame 3D rotary attention mechanism for antibody-specific epitope prediction.
By expressing pairwise residue displacements in backbone-defined local frames, LFRA brings folded 3D structural geometry into query--key interactions while preserving invariance to SE(3) transformations.
Experiments on the AsEP benchmark, including comparisons with 1D rotary attention, global-frame variants, and alternative geometric modules, demonstrate that LFRA improves antibody-conditioned epitope prediction, with particularly notable gains in MCC and F1.
Additional transformation and coordinate noise analyses validate the geometric properties of LFRA and demonstrate its stability under global rigid transformations and residue-level coordinate perturbations.
Finally, antibody ablation experiments confirm that LFRA relies on antibody-specific information and the correct antibody--antigen pairing, rather than merely learning generic antigen-surface binding preferences.
Together, these findings establish local-frame rotary attention as an effective approach to integrating protein structure into antibody-specific epitope prediction and motivate further exploration of geometry-aware representations for protein interaction modeling.

\bibliographystyle{splncs04}
\bibliography{main}

\clearpage
\appendix

\section{Rotary Frequency Design}
\label{sec:ladder}
We determine the frequency range from reference structural scales and the number of frequencies from the available attention head dimensions.
For a scalar displacement $u$, the phase pair $(\cos\omega u,\sin\omega u)$ has period $2\pi/\omega$.
To preserve distinguishability over a reference spatial extent $D_{\mathrm{ref}}$, we require the phase to vary by less than half a cycle over $[-D_{\mathrm{ref}},D_{\mathrm{ref}}]$.
Conversely, to avoid excessive phase variation at typical residue spacing $\ell_{\mathrm{ref}}$, we limit the phase change over this distance to at most $\pi$.
These considerations yield the following frequency constraints:
\begin{equation}
\omega_{\min}D_{\mathrm{ref}}<\pi,
\qquad
\omega_{\max}\ell_{\mathrm{ref}}\leq\pi.
\label{eq:frequency_scale_conditions}
\end{equation}

We estimate these reference scales from the training data.
Across all $1{,}383$ training complexes in each of the two official AsEP splits, the largest $\mathrm{C}_{\alpha}$ separation between modeled residues within either partner is $D_{\mathrm{ref}}=204.43$~\AA{}, while the median nearest-neighbor distance is $\ell_{\mathrm{ref}}=3.80$~\AA{} in both splits.
These statistics are computed on antibody CDR and antigen surface residues with resolved backbone coordinates, pooling per-residue nearest-neighbor distances within each training split.
The resulting endpoint scales are approximately $0.0154$ and $0.827$~$\mathrm{rad}\,\text{\AA}^{-1}$.
We therefore use the following rounded frequency bounds:
\begin{equation}
\omega_{\min}=0.015~\mathrm{rad}\,\text{\AA}^{-1},\qquad
\omega_{\max}=0.8~\mathrm{rad}\,\text{\AA}^{-1}.
\label{eq:frequency_endpoints}
\end{equation}
The slowest frequency provides an injective phase window of $|u|<\pi/\omega_{\min}\approx209.44$~\AA{}, while the fastest frequency produces a phase change of $3.04$ radians over the median nearest-neighbor distance.

Each frequency requires one two-dimensional rotation plane per local axis, giving the channel constraint
\begin{equation}
6F\leq d_h,\qquad
F=\left\lfloor\frac{d_h}{6}\right\rfloor.
\label{eq:frequency_budget}
\end{equation}
We use the largest feasible $F$ to maximize frequency coverage at a fixed head dimension.
For $F\geq2$, the frequencies are uniformly spaced on a logarithmic scale:
\begin{equation}
\omega_f=\omega_{\min}
\left(\frac{\omega_{\max}}{\omega_{\min}}\right)^{\frac{f-1}{F-1}},
\qquad f=1,\ldots,F.
\label{eq:frequency_ladder}
\end{equation}
With $d=256$ and $H=8$, each head has $d_h=32$, allowing $F=5$ frequencies and leaving $r=2$ unrotated channels.
The resulting frequency set is approximately $\{0.015,0.0405,0.1095,0.2960,0.8000\}$~$\mathrm{rad}\,\text{\AA}^{-1}$.
These frequencies are shared across local axes, attention heads, and partners, and remain fixed during training.

\section{Details}
\label{sec:appendix_implementation_details}

Table~\ref{tab:implementation_hyperparameters} summarizes the architecture, training configuration, and evaluation protocols for LFRA.
Model selection and threshold calibration use only the validation set: we select the checkpoint with the highest average precision and determine the classification threshold by maximizing validation MCC.
Test metrics are computed over pooled antigen surface residues.
Rigid-transformation analyses are performed exclusively at inference, with model parameters and validation-selected thresholds held fixed.
Within each split, identical sampled test subsets and transformations are used across methods and training seeds to ensure consistent comparisons.
AUCROC is not reported for EpiPred, ESMFold, or MaSIF-site because their outputs do not provide residue-level continuous scores directly comparable to those of the other methods. For EpiPred and ESMFold, interface residues are derived from the predicted structures using a geometric proximity criterion, resulting in binary residue predictions. MaSIF-site assigns probabilities to surface-mesh vertices rather than directly to residues; residues are subsequently labeled as epitopes when they are close to vertices whose predicted probability exceeds (0.7), likewise producing thresholded binary predictions. We therefore report only threshold-based evaluation metrics for these methods~\cite{NEURIPS2024_15add673}.

\begin{table*}[ht]
\centering
\caption{Model architecture, training, and evaluation configurations for LFRA.}
\small
\begin{tabular}{ll}
\toprule
Hyperparameter & Setting \\
\midrule
\multicolumn{2}{l}{\textit{Model Architecture}} \\
Embedding dimensions (antibody / antigen) & 512 / 480 \\
ESM-2 checkpoint & \texttt{esm2\_t12\_35M\_UR50D} \\
AntiBERTy checkpoint & \texttt{AntiBERTy\_md\_smooth} (via IgFold) \\
Encoder depth & 1 layer per partner; separate parameters \\
Cross-attention depth & 2 layers \\
Hidden width / attention heads & 256 / 8 \\
Head dimension & 32 \\
Feed-forward width / activation & 512 / GELU \\
Attention-block normalization & Pre-layer normalization \\
Dropout & 0.1 \\
Prediction-head dimensions & $768\rightarrow256\rightarrow1$ \\
\midrule
\multicolumn{2}{l}{\textit{Geometric Encoding}} \\
Rotary frequencies & 5 per axis; logarithmically spaced \\
Rotary frequency range & $0.015$--$0.8$~rad\,\AA$^{-1}$ \\
Frequency configuration & Fixed across axes, heads, and partners \\
Rotary / non-rotary dimensions & 30 / 2 per head \\
\midrule
\multicolumn{2}{l}{\textit{Training}} \\
Loss & Weighted BCE $+\,0.1\times$ soft Dice \\
Dice smoothing constant & 1 \\
Positive-class weight & Training-set non-epitope / epitope residue ratio \\
Optimizer & AdamW \\
Learning rate / weight decay & $10^{-4}$ / $10^{-5}$ \\
Batch size / maximum epochs & 1 complex / 100 \\
Gradient-norm clipping & 1.0 \\
Learning-rate scheduler & ReduceLROnPlateau; factor 0.5, patience 5 \\
Checkpoint selection & Highest validation AUPRC \\
Early stopping & 15 epochs without validation improvement \\
Random seeds & 42, 43, 44, 45, 46 \\
\midrule
\multicolumn{2}{l}{\textit{Evaluation}} \\
AUPRC & Average precision over pooled antigen surface residues \\
Decision threshold & Maximum validation MCC; fixed for test evaluation \\
Result reporting & Mean $\pm$ sample standard deviation across five training seeds \\
\midrule
\multicolumn{2}{l}{\textit{Transformation Analysis}} \\
Whole-complex subset & 50 test complexes per split; sampling seed 42 \\
Independent Ag/Ab subsets & Two draws of 50 test complexes per split; seeds 42 and 43 \\
Whole-complex transformation & Seed 42; same rigid transform applied to both partners \\
Independent transformation & Seeds 42 and 43; separately sampled transforms for Ag/Ab \\
Transformation types & Translation, random $\mathrm{SO}(3)$ rotation, and $\mathrm{SE}(3)$ transformation \\
Rotation sampling & Uniform random unit quaternions \\
Translation sampling & Random direction; magnitude uniform in $[0,D]$ \\
Translation scale $D$ & Diameter of selected, coordinate-valid $\mathrm{C}_{\alpha}$ atoms from both Ag/Ab \\
Model seeds & 42, 43, 44, 45, 46 \\
\bottomrule
\end{tabular}
\label{tab:implementation_hyperparameters}
\end{table*}

\end{document}